\pdfoutput=1

\documentclass[11pt]{article}

\usepackage[]{acl}

\usepackage{times}
\usepackage{latexsym}

\usepackage{graphicx}

\usepackage[justification=centering]{caption}
\usepackage{subcaption}

\usepackage{multirow}

\usepackage{float}

\usepackage[T1]{fontenc}

\usepackage[utf8]{inputenc}

\usepackage{microtype}

\title{Investigating Stylistic Profiles for the Task of Empathy Classification in Medical Narrative Essays}

\author{Priyanka Dey \\
  Computer Science Department \\
  University of Illinois, Urbana-Champaign \\
  \texttt{pdey3@illinois.edu} \\\And
  Roxana Girju \\
  Department of Linguistics,\\
  Computer Science Department, \\
  Beckman Institute, \\
  University of Illinois, Urbana-Champaign \\
  \texttt{girju@illinois.edu} \\}

\begin{document}
\maketitle
\begin{abstract}
One important aspect of language is how speakers generate utterances and texts to convey their intended meanings. 
In this paper, we bring various aspects of the Construction Grammar (CxG) and the Systemic Functional Grammar (SFG) theories in a deep learning computational framework to model empathic language. 
Our corpus consists of 440 essays written by premed students as narrated simulated patient–doctor interactions. 
We start with baseline classifiers (state-of-the-art recurrent neural networks and transformer models). Then, we enrich these models with a set of linguistic constructions proving the importance of this novel approach to the task of empathy classification for this dataset. Our results indicate the potential of such constructions to contribute to the overall empathy profile of first-person narrative essays. 

\end{abstract}


\section{Introduction}

Much of our everyday experience is shaped and defined by actions and events, thoughts and perceptions which can be accounted for in different ways in the system of language. The grammatical choices we make  
when writing an essay (i.e., pronoun use, active or passive verb phrases, sentence construction) differ from those we use to email someone, or those we utter in a keynote speech. 
"Word choice and sentence structure are an expression of the way we attend to the words of others, the way we position ourselves in relation to others" \cite{micciche2004making}. Such choices allow us to compare not only the various options available in the grammar, but also what is expressed in discourse with what is suppressed \cite{menendez2017christopher}.

Given the great variability in the modes of expression of languages, the search for an adequate design of grammar has long motivated research in linguistic theory.
One such approach is CxG \cite{paul1999grammatical,goodberg1995constructions,fillmore2006construction} which prioritizes the role of constructions, conventional form-meaning pairs, in the continuum between lexis and syntax \cite{goldberg2009constructions}.
As such, these constructions form a structured inventory of speakers’ knowledge of the conventions of their language \cite{langacker1987foundations}.

Another particular grammatical facility for capturing experience in language is Halliday’s system of transitivity as part of the Systemic Functional Grammar (SFG) \cite{halliday1994introduction,halliday2014introduction}, a theory of language centred around the notion of language function. SFG pays great attention to how speakers generate utterances and texts to convey their intended meanings. This can make our writing effective, but also give the audience a sense of our own personality. However, unlike CxG, Halliday’s system of transitivity describes the way in which the world of our experience is divided by grammar into a ‘manageable set of process types’ \cite{halliday2014introduction} each offering not only a form-meaning mapping, but also a range of stylistic options for the construal of any given experience through language. 
In stylistics, researchers have used this model to uncover and study the grammatical patterns through which texts can enact a particular ideology, or an individual’s distinctive ‘mind style’ of language \cite{fowler1996}.

The idea of ‘style as choice’ in Halliday’s transitivity system can be best understood as experiential strategies (like avoiding material processes or repeating passive voice constructions) such as those identified as contributing to a reduced sense of awareness, intentionality or control in the human agent responsible \cite{fowler2013linguistics,simpson2014action}. Such an individual is often said to appear ‘helpless’ and ‘detached’ \cite{halliday2019linguistic,simpson2003language}, or  ‘disembodied’ \cite{hoover2004altered}. Take for instance, construction choices like 'I reassured her' vs. 'She was reassured', or "I greeted her upon entrance" vs. "The nurse greeted her upon entrance" vs. "She was greeted upon entrance" -- which show the degree of agency and intended involvement on the part of the agent in the action.
Such linguistic choices often occur together in stylistic profiling exercises to showcase the techniques contributing to ‘passivity’, or the degree of suppression of agency and power in characterisation \cite{kies1992uses}. 

In this paper, we try to bring CxG and SFG closer together in the study of discourse level construction of arguments for the analysis of empathic content of narrative essays. Specifically, inspired by research in critical discourse analysis, we are taking  a step further to show ways in which such construction choices can manipulate (and even reduce) the attention we give to the agency and moral responsibility of individuals \cite{jeffries2017critical,van2017discourse}. 
Specifically, 
such form-meaning-style mappings can be used to capture the point of view as an aspect of narrative organization and the perspective through which a story is told, the way the characters are portrayed in terms of their understanding of the processes they are involved in, as well as their own participation in the story. 
In this respect, "narratives seem necessary for empathy [..] they give us access to contexts that are broader than our own contexts and that allow us to understand a broad variety of situations" \cite{Gallagher2012}.
They provide a form/structure that allows us to frame an understanding of others, together with a learned set of skills and practical knowledge that shapes our understanding of what we and others are experiencing. 

Drawing on Halliday’s transitivity framework rooted in Systemic Functional Linguistics, this paper attempts to reveal the (dis)engaged style of empathic student essays from a semantic-grammatical point of view. 
Specifically, we want to investigate how certain types of processes (i.e., verbs) and constructions (i.e., passive voice) function to cast the essay writers (as main protagonists and agents) as perhaps rather ineffectual, passive, and detached observers of the events around them and of the patient's emotional states.

We take a narrative approach to empathy and explore the experiences of premed students at a large university by analysing their self-reflective writing portfolios consisting of a corpus of first-person essays written by them as narrated simulated patient-doctor interactions. 
The corpus has been previously annotated and organized \cite{Shi-etal2021,michalski2022empathy} following established practices and theoretical conceptualizations in psychology (Cuff et al., 2016; Eisenberg et al., 2006; Rameson et al., 2012). Computationally, we introduce a set of informative baseline experiments using state-of-the-art recurrent neural networks and transformer models for classifying the various forms of empathy. As initial experiments show relatively low scores, we measure the presence of several grammatical structures, 
leveraging Halliday’s theory of transitivity, and its correlation with the essays’ overall empathy scores. We apply this framework to state-of- the-art and representative neural network models and show significant improvement in the empathy classification task for this dataset. Although previous research suggests that narrative-based interventions tend to be effective education-based methods, it is less clear what are some of the linguistic mechanisms through which narratives achieve such an effect, especially applied to empathy, which is another contribution of this research.

\section{Related Work}
In spite of its increasing theoretical and practical interest,
empathy research in computational linguistics has been relatively sparse and limited to empathy recognition, empathetic response generation, or empathic language analysis in counselling sessions. 
Investigations of empathy as it relates to clinical practice have received even less attention given the inherent data and privacy concerns. 

Most of the research on empathy detection has focused on spoken conversations or interactions, some in online platforms (e.g. 
\cite{perez-etal2017,khanpour2017,otterbacher-etal2017,sharma-etal2021,hosseini-caragea-2021-distilling-knowledge}, very little on narrative genre \cite{buechel2018modeling,wambsganss2021supporting}, and even less in clinical settings. \citet{buechel2018modeling} used crowd-sourced workers to self-report their empathy and distress levels and to write empathic reactions to news stories. \citet{wambsganss2021supporting} built a text corpus of student peer reviews collected from a German business innovation class annotated for cognitive and affective empathy levels. 
Using Batson’s Empathic Concern-Personal Distress Scale \cite{batson1987distress}, \citet{buechel2018modeling} have focused only on negative empathy instances (i.e., pain and sadness "by witnessing another person’s suffering"). However, empathy is not always negative \cite{Fan-etal2011}. A dataset reflecting empahatic language should ideally allow for expressions of empathy that encompass a variety of emotions, and even distinguish between sympathy and empathy.\footnote{ 
Some studies don't seem to differentiate between sympathy and empathy \cite{rashkin2018towards,lin2019moel}.}

Following a multimodal approach to empathy prediction, \citet{Frankel2000} and \citet{Cordella-Musgrave2009} identify sequential patterns of empathy in video-recorded exchanges between medical graduates and cancer patients. \citet{sharma-etal-2020-computational} analyzed the discourse of conversations in online peer-to-peer support platforms. Novice writers were trained to improve low-empathy responses and provided writers with adequate feedback on how to recognize and interpret others’ feelings or experiences. In follow-up research, they performed a set of experiments \cite{sharma-etal2021} whose results seemed to indicate that empathic written discourse should be coherent, specific to the conversation at hand, and lexically diverse. 

To our knowledge, no previous research has investigated the contribution of grammatical constructions like Halliday's transitivity system to the task of empathy detection in any genre, let alone in clinical education.\footnote{Besides our own research \cite{Shi-etal2021,michalski2022empathy,dey-girju-louhi2022,girju-girju-2022-design}.}

\section {Self-reflective Narrative Essays in Medical Training}

Simulation‐based education (SBE) is an important and accepted practice of teaching, educating, training, and coaching health-care professionals in simulated  environments \cite{bearman2019power}. Four decades-worth of SBE research has shown that “simulation technology, used under the right conditions … can have large and sustained effects on knowledge and skill acquisition and maintenance among medical learners” \cite{mcgaghie2014critical}. In fact, simulation‐based education, an umbrella term that covers a very broad spectrum of learning activities from communication skill role‐playing to teamwork simulations, is known to contribute to shaping experiences in undergraduate and postgraduate medical, nursing and other health education. In all these activities, learners contextually enact a task which evokes a real‐world situation allowing them to undertake it as if it were real, even though they know it is not \cite{dieckmann2007deepening,bearman2003virtual}.

Personal narratives and storytelling can be viewed as central to social existence \cite{bruner1991narrative}, as stories of lived experience \cite{van2016researching}, or as a way in which one constructs notions of self \cite{ezzy1998theorizing}.
In this research, we focus on self-reflective narratives written by premed students given a simulated scenario. Simulation is strongly based on our first-person experiences since it relies on resources that are available to the simulator. In a simulation process, the writer puts themselves in the other’s situation and asks "what would I do if I were in that situation?” 
Perspective taking is crucial for fostering affective abilities, enabling writers to imagine and learn about the emotions of others and to share them, too. As empathy is other-directed \cite{de2012like,Gallagher2012}, this means that we, as narrators, are open to the experience and the life of the other, in their context, as we can understand it.
Some evidence shows that we can take such reliance on narrative resources to open up the process toward a more enriched and non-simulationist narrative practice (i.e., real doctor-patients interactions in clinical context) \cite{Gallagher2012}.

This study's intervention was designed as a written assignment in which premed students were asked to consider a hypothetical scenario where they took the role of a physician breaking the news of an unfavorable diagnosis of high blood cholesterol to a middle-aged patient\footnote{The patient was referred to as Betty, initially. Later in the data collection, students could also identify the patient as John.}. They were instructed to recount (using first person voice) the hypothetical doctor-patient interaction where they explained the diagnosis and prescribed medical treatment to the patient using layman terms and language they believed would comfort as well as persuade the hypothetical patient to adhere to their prescription. 
Prior to writing, students completed a standard empathic training reading assignment \cite{baile2000spikes}. 
They received the following prompt instructions and scenario information.\footnote{All data collected for this study adheres to the approved Institutional Review Board protocol.}

\underline{Prompt Instructions:} Imagine yourself as a physician breaking bad news to a patient. Describe the dialogue between the patient and you, as their primary care physician. In your own words, write an essay reporting your recollection of the interaction as it happened (write in past tense). Think of how you would break this news if you were in this scenario in real life. 
In your essay, you should be reflecting on (1) how the patient felt during this scenario and (2) how you responded to your patient's questions in the scenario below. 

\underline{Scenario:} Betty is 32 years old, has a spouse, and two young children (age 3 and 5). You became Betty’s general practitioner last year. Betty has no family history of heart disease. In the past 6 months, she has begun experiencing left-side chest pain. Betty’s bloodwork has revealed that her cholesterol is dangerously high. Betty will require statin therapy and may benefit from a healthier diet and exercise.

With the students' consent, we collected a corpus of 774 essays over a period of one academic year \cite{Shi-etal2021}. Following a thorough annotation process, annotators (undergraduate and graduate students in psychology and social work)\footnote{The students were hired based on previous experience with similar projects in social work and psychology.} labeled a subset of 440 randomly selected essays at sentences level following established practices in psychology \cite{cuff2016empathy,eisenberg2006prosocial,rameson2012neural}. 
The labels are: \emph{cognitive empathy} (the drive and ability to identify and understand another’s emotional or mental states; e.g., "She looked tired"); \emph{affective empathy} (the capacity to experience an appropriate emotion in response to another’s emotional or mental state; e.g.: "I felt the pain"); and \emph{prosocial behavior} (a response to having identified the perspective of another with the intention of acting upon the other’s mental and/or emotional state; e.g.: "I reassured her this was the best way"). 
Everything else was "no empathy". 
The six paid undergraduate students were trained on the task and instructed to annotate the data. 
Two meta-annotators, paid graduate students with prior experience with the task, reviewed the work of the annotators and updated the annotation guidelines at regular intervals, in an iterative loop process after each batch of essays\footnote{10 essays per week}.
The meta-annotators reached 
a Cohen’s kappa of 0.82, a good level of agreement. Disagreed cases were discussed and mitigated. At the end, all the essays were re-annotated per the most up-to-date guidelines. 

In this paper, we collapsed all the affective, cognitive, and prosocial empathy labels into one \emph{Empathy Language} label -- since we are interested here only in emphatic vs. non-empathic sentences. After integrating the annotations and storing the data for efficient search \cite{michalski2022empathy}, our corpus consisted of 10,120 data points (i.e., sentences) highlighted or not with empathy. Each essay was also rated by our annotators with a score on a scale from 1-5 (one being the lowest) to reflect overall empathy content at essay level.  


\section{Constructions and Stylistic Profiles in Empathic Narrative Essays}

In CxG, constructions can vary in size and
complexity -- i.e., morphemes, words, idioms, phrases, sentences. 
In this paper, we focus mainly on simple sentence-level constructions\footnote{We also consider constructions at word level - i.e., verbs.}, which, since we work with English, are typically of the form S V [O], where S is the subject, V is the verb, and O is the object (e.g., a thing, a location, an attribute).
For instance, "Betty took my hand" matches the construction S V O with the semantics <Agent Predicate Goal>.
SFG and CxG give the same semantic analysis, modulo some
terminological differences \cite{lin2006systemic}. Specifically, they agree that the sentence above describes a process (or a predicate), which involves two
participant roles providing the same linking relationship between the
semantic and the syntactic structures: an Actor (or Agent) / Subject, and a Goal (Patient) / Object. 

We start by checking whether the subject of a sentence consists of a human or a non-human agent. After identifying the grammatical subjects in the dataset's sentences with the Python Spacy package, we manually checked the list of human agents (the five most frequent being \textit{I} (24.56\%), \textit{She} (5.76\%), 
\textit{Betty} (18.43\%), 
\textit{John} (6.24\%), 
\textit{Patient} (4.86\%)).\footnote{Other subjects: \textit{Nurse}, \textit{Doctor}, \textit{Family}, \textit{Children}, \textit{Wife}, \textit{Husband}, and \textit{Spouse}}

Halliday's transitivity model describes the way in which the world of our experience can be divided by grammar into a manageable set of process types, the most basic of which are: \emph{material processes} (external actions or events in the world around us; e.g., verbs like "write", "walk", "kick") and \emph{mental processes} (internal events; e.g., verbs of thinking, feeling, perceiving). We first identify sentences containing material  
and mental processes  
by extracting the verbs in each sentence
(Table \ref{tab:hall}). 
About 75\% of the dataset contains such processes, with
material processes appearing more frequently than mental ones (by a small margin: 0.9\%).

Inspired by the success of Halliday's transitivity system on cognitive effects of linguistic constructions in literary texts \cite{nuttall2019transitivity}, we also examine a set of construction choices which seem to co-occur in texts 
as material and mental actions or events. 
In our quest of understanding empathy expression in student narrative essays, we want to test if such contributions lead to a reduced sense of intentionality, awareness or control for the agentive individual represented (i.e., the essay writer in the role of the doctor), and thus, identifying the stylistic profile of the narrative.
Specifically, these constructions are: \emph{Human Actor + Process (HA+P); Body Part + Process (BP+P); Other Inanimate Actor + Process (IA+P); Goal + Process (G+P)} (see Table \ref{tab:hall}). We identify HA+P to be the most common construction within our dataset, appearing in just less than half of the sentences (49.82\%). The remaining constructions are much rarer with G+P being the least frequent (12.54\%).   

Drawing from \cite{langacker1987foundations}, \citet{nuttall2019transitivity} also notes that these experiences can vary in force-dynamic (energetic) quality and thus sentences exhibiting an energetic tone are linked with ‘high’ transitivity and those with lower or static energy can be linked to ‘low’ transitivity. In order to identify energetic sentences, we leverage the IBM Watson Tone Analyzer API \cite{ibm} which assesses the emotions, social propensities, and language styles of a sentence. We denote sentences containing high extroversion and high confidence (values > 0.8) as energetic. Sentences with low scores are marked as static. 61.77\% of the sentences exhibit a static tone, energetic tone being less frequent. 
 
\begin{figure}
    \centering
    \small
    \includegraphics[scale=0.35]{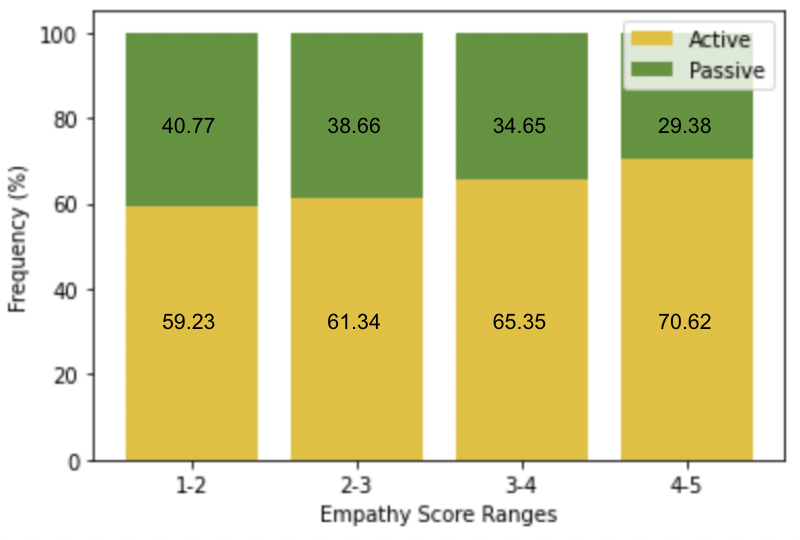}
    \caption{Frequency distribution (\%) of voice in essays for various overall empathy score ranges}
    \label{fig:voice}
\end{figure}

In SFG, active and passive voice plays an important role as well. \citet{nuttall2019transitivity} shows that, in some genres, text indicating a lower degree of agentive control tends to use more passive voice constructions. As this is also relevant to our task, we test whether voice  contributes indeed to a reduced sense of intentionality, awareness or control for the Agent (in particular the essay writer playing the doctor's role) and how these features correlate with the overall empathy score at  essay level. Using an in-house grammatical-role extraction tool developed on top of Spacy's dependency parser, we find that 66\% of sentences use active voice and 34\% passive voice.\footnote{The active/passive voice ratio varies per genre \cite{strunk2007elements}. Note that in a sentence using passive voice, the subject is acted upon, which shows the main character's degree of detachment, which is of interest here.} 
77.92\% of active-voice sentences exhibit human actor subjects and only 22.08\% include non-human actors. Similarly for passive voice, the majority (83.09\%) of sentences had human actors.
Comparing frequencies of active and passive voice across various essay empathy score ranges (Figure \ref{fig:voice}), we notice that higher empathy essays (scores >3) seem to rely more on active voice (65-70\% of the sentences in active voice) as opposed to lower empathy essays (scores < 3) which have less than 65\% of sentences in active voice.

\begin{table*}[ht]
    \centering
    \small
    \begin{tabular}{p{1.43cm} p{1.5cm} p{5cm} p{6cm}} \hline
         \textbf{Feature} & \textbf{Frequency} & \textbf{Definition} & \textbf{Example} \\ \hline
         \textit{Active} & 62.12\% & the subject of the sentence is the one doing the action expressed by the verb & "I watched as the patient slowly sat down in the chair." \\ 
         \textit{Passive} & 37.88\% & the subject is the person or thing acted on or affected by the verb's action & "The patient I just had an appointment with is named Betty." \\ 
         \textit{Material} & 37.39\% &  external actions or events in the world around us & "The nurse had already retrieved the bloodwork reports and handed them to me before I entered the room." \\ 
         \textit{Mental} & 36.49\% & events/feelings expressed by a user & " 'I can imagine that you have several questions, so I am happy to answer any questions or clear any doubts you might have.' I said to her. " \\ 
         \textit{HA+P} & 49.82\% & consists of a human actor and a material/mental process &  "I calmly started explaining the treatment options." \\ 
         \textit{BP+P} & 15.85\% & consists of a non-human actor related to body parts in material/mental process & "Her shoulders started shaking when she heard the news, and I could tell she would need some time to process the news." \\
         \textit{IE+P} & 18.34\% & consists of an inanimate actor in material/mental process & "The file was already in the room when I arrived." \\
         \textit{G+P} & 12.54\% & consists of the passivisation of material/mental process and deletion of actor & "The effects of her lifestyle had already started to affect her physical strength." \\
         \textit{Energetic} & 38.23\% & e.g., high extroversion and confidence & "I could see Betty fidgeting with her fingers as she began to process the news." \\
         \textit{Static} & 61.77\% & e.g., low extroversion and  confidence & "The nurse brought in the file quickly." \\
          \hline
         
    \end{tabular}%
    \caption{Our set of SFG's transitivity constructions with their distribution and examples. Note that the total distribution should not add to 100\%, as these are not mutually exclusive features.}
    \label{tab:hall}
\end{table*}

Stylistic research has also shown \cite{nuttall2019transitivity} the importance of movement of body parts as non-human agents. 
We, too, parsed sentences for the use of body parts, i.e. \textit{eyes}, \textit{arms}, \textit{head} and curated a list based on anatomical terminology as defined by \citet{body_parts} resulting in about 18.61\% of the dataset sentences (statistics for top 5 most common bodyparts are in Table \ref{tab:body_parts}).

Table \ref{tab:hall} summarize all the identified constructions and stylistic features discussed in this section.

\begin{table*}
    \centering
    \small
    \begin{tabular}{p{1.65cm} p{4cm} p{1.65cm} p{6.5cm}} \hline
         \textbf{Body Part} & \textbf{POS Used} & \textbf{Frequency} & \textbf{Example} \\ \hline
         
         \textit{Eye} & subject, indirect object, prepositional object & 42.96\% & "I saw in her eyes tears forming as she realized the gravity of the issue at hand." \\

         \textit{Hand} & subject, prepositional object, indirect object, direct object & 16.14\% & "John began clasping his hands." \\

         \textit{Head} & direct object, indirect object & 8.60\% & "John shook his head as he sat down across from me." \\

         \textit{Shoulder} & subject, prepositional object, direct object & 5.47\% & "The patient shrugged his shoulders." \\

         \textit{Body} & subject, prepositional object, direct object & 4.99\% & "The vitals showed that the patient’s body was not in its healthiest form." \\
         
          \hline
         
    \end{tabular}%
    \caption{Most common body parts in the empathy essay dataset}
    \label{tab:body_parts}
\end{table*}

\section{Empathy Classification Task}

Our ultimate goal is to build an informed and performant classifier able to determine the degree of empathetic content of a medical essay overall and at sentence level.
Taking advantage of form-meaning-style mappings in the language system, in this paper, we built and test a number of state-of-the-art classifiers enriched with varied constructions and stylistic features (Table \ref{tab:hall}) which are described next. 

\subsection{Identification of Sentence Themes}

In medical training, students learn not only how to diagnose and treat patients’ medical conditions, but also how to witness the patient's illness experience. In fact, in practical interactions with patients, they often switch between these positions: empathizing with the patient's situation (i.e., witnessing what it is like for the patient), and providing medical care (i.e., understanding what they need medically). 

As such, we wanted to capture the distribution of such emphatic content and medical information in our narrative essays of hypothetical doctor-patient interactions. Specifically, we looked at recurring topics within sentences and identified the following themes in our dataset at the sentence level: \textit{Medical Procedural Information; Empathetic Language; Both} (Medical and Empathetic Language); and \textit{Neither}. Sentences referring to \textit{Medical Procedural Information} were identified based on keyword matching following established medical term vocabulary generated from Dr. Kavita Ganesan’s work on clinical concepts \cite{Ganesan2016}. Sentences containing \textit{Empathetic Language} were already annotated manually by our annotators for each essay at the sentence level (see Section 3).
Sentences containing both medical procedural info and empathetic content were marked as \textit{Both}, while remaining sentences are marked as \textit{Neither}. Table \ref{tab:themes} shows these categories, their definitions, examples and counts per category (10,120 sentences overall). We also give examples of two essays highlighted with these themes in the Appendix (Section \ref{sec:appendix}).

In the next sections we present the classification results of various multi-class machine learning models (for each of the 4 themes: \textit{Medical Procedural Information}, \textit{Empathetic Language}, \textit{Both}, and \textit{Neither}). 

\begin{table*}
    \centering
    \small
    \begin{tabular}{p{4cm}p{0.8cm} p{9.8cm}} \hline
         \textbf{Theme} & \textbf{Freq.} & \textbf{Example} \\ \hline
         \textit{Medical Procedural Information} & 37.39\% & "The patient’s vitals showed that his body was not healthy and it was necessary to make some diet and lifestyle changes." \\ 
         \textit{Empathetic Language} & 36.49\% & "I noticed Betty looked confused and so I tried to reassure her we would do everything possible to make the changes in her lifestyle." \\ 
         \textit{Both} & 21.28\% &  "I knew the statin treatment could be difficult, so I wanted to make sure Betty felt comfortable and understood the procedure." \\ 
         \textit{Neither} & 4.84\% & "The file was left on the counter, and I picked it up before going in to see Betty." \\
         
          \hline
         
    \end{tabular}%
    \caption{Examples and distribution of identified themes in sentences}
    \label{tab:themes}
\end{table*}

         
         

\begin{table*}[t]
    \centering
    \resizebox{\textwidth}{!}{
     \begin{tabular}{lccc|ccc|ccc|ccc} \hline
        \multirow{2}{*}{\textbf{Classifier}} & \multicolumn{3}{c}{\textbf{Medical Procedural Information}} & \multicolumn{3}{c}{\textbf{Empathetic Language}} & 
        \multicolumn{3}{c}{\textbf{Both}} &
        \multicolumn{3}{c}{\textbf{Neither}} 
        \\ \cline{2-4} \cline{5-7} \cline{8-10} \cline{11-13}
        & Prec. & Rec. & F1 & Prec. & Rec. & F1 & Prec. & Rec. & F1 & Prec. & Rec. & F1 \\ \hline 
        SVM & 0.70 & 0.68 & 0.69 & 0.52 & 0.61 & 0.56 & 0.49 & 0.47 & 0.48 & 0.78 & 0.39 & 0.51 \\ 
        LogR & 0.62 & 0.67 & 0.64 & 0.49 & 0.54 & 0.51 & 0.51 & 0.53 & 0.52 & 0.68 & 0.61 & 0.64  \\ 
        LSTM  & 0.64 & 0.69 & 0.67 & 0.51 & 0.54 & 0.52 & 0.59 & 0.53 & 0.56 & 0.66 & 0.61 & 0.63  \\ 
        biLSTM & 0.65 & 0.7 & 0.68 & 0.51 & 0.54 & 0.52 & 0.56 & 0.53 & 0.54 & 0.68 & 0.62 & 0.65 \\ 
        CNN & 0.70 & 0.71 & 0.70 & 0.52 & 0.54 & 0.53 & 0.64 & 0.53 & 0.57 & 0.71 & 0.63 & 0.66 \\  
        BERT & 0.69 & 0.72 & 0.70 & 0.55 & 0.61 & 0.58 & 0.57 & 0.63 & 0.60 & 0.68 & 0.65 & 0.66 \\ \hline \hline 
        constructionBERT & 0.71 & 0.73 & 0.72 & 0.64 & 0.67 & 0.65 & 0.76 & 0.58 & 0.66 & 0.78 & 0.72 & 0.75 \\ 
        constructionBERT-\textit{Voice:Active} & 0.71 & 0.73 & 0.72 & 0.58 & 0.63 & 0.65 & 0.64 & 0.64 & 0.62 & 0.77 & 0.72 & 0.74 \\ 
        constructionBERT-\textit{Voice:Passive} & 0.71 & 0.73 & 0.72 & 0.65 & 0.67 & 0.66 & 0.76 & 0.61 & 0.67 & 0.78 & 0.72 & 0.75 \\ 
        constructionBERT-\textit{Process:Material} & 0.70 & 0.72 & 0.71 & 0.61 & 0.65 & 0.63 & 0.68 & 0.58 & 0.63 & 0.78 & 0.72 & 0.75 \\ 
        constructionBERT-\textit{Process:Mental} & 0.70 & 0.72 & 0.71 & 0.59 & 0.63 & 0.61 & 0.66 & 0.58 & 0.62 & 0.78 & 0.71 & 0.74 \\ 
        constructionBERT-\textit{HA+P} & 0.69 & 0.72 & 0.70 & 0.59 & 0.64 & 0.62 & 0.66 & 0.58 & 0.62 & 0.68 & 0.69 & 0.68 \\ 
        constructionBERT-\textit{BP+P} & 0.71 & 0.73 & 0.72 & 0.55 & 0.64 & 0.59 & 0.61 & 0.63 & 0.62 &  0.71 & 0.72 & 0.71 \\ 
        constructionBERT-\textit{IE+P} & 0.70 & 0.73 & 0.71 & 0.61 & 0.64 & 0.62 & 0.73 & 0.57 & 0.64 & 0.76 & 0.72 & 0.74 \\ 
        constructionBERT-\textit{G+P} & 0.71 & 0.73 & 0.72 & 0.64 & 0.66 & 0.65 & 0.74 & 0.56 & 0.64 & 0.78 & 0.72 & 0.75  \\
        constructionBERT-\textit{Tone:Energetic} & 0.71 & 0.73 & 0.72 & 0.58 & 0.62 & 0.60 & 0.66 & 0.57 & 0.61 & 0.78 & 0.72 & 0.75 \\ 
        constructionBERT-\textit{Tone:Static} & 0.71 & 0.73 & 0.72 & 0.64 & 0.62 & 0.63 & 0.71 & 0.58 & 0.64 & 0.78 & 0.73 & 0.75 \\ 
         
         \hline
    \end{tabular}
    }
    \caption{Precision, recall and F1 scores of all baseline classifiers on the imbalanced test dataset: 770 \textit{Medical Procedural Information}, 722 \textit{Empathetic Language}, 433 \textit{Both}, 98 \textit{Neither} sentences}
    \label{tab:themes_scores}
\end{table*}

\subsection{Baseline Models and Analysis}

In evaluating several state-of-the-art machine learning algorithms, we started with two representative baseline models: support vector machines (SVM) and logistic regression (logR). As we are interested in observing the performance of deep learning methods, we also experiment with long-short term memory (LSTM) \cite{hochreiter1997long}, bidirectional long-short term memory (bi-LSTM) \cite{graves2005framewise}, and convolutional neural network (CNN)  \cite{cnn} models; additionally, we use the transformer models BERT \cite{devlin2018bert} and roBERTa. 

As we are performing sentence classification, our features are unigrams (single words). For the logistic regression models, we used a L2 regularization and for the SVM models, a linear kernel function. We initialized the embedding layers in our neural models (LSTM, bi-LSTM, CNN) with GloVe embeddings since the expression of empathy involves larger units than words, and embeddings are known to better capture contextual information. We further decided to apply an attention layer to these models to learn patterns that may improve the classification. For the transformer BERT and roBERTa models, we use the default embeddings and apply a dropout layer with probability 0.4 which helps to regularize the model; we use a linear output layer and apply a sigmoid on the outputs. 
For each type of theme, we reserve an 80/20 training/test ratio, with 5-fold cross validation.
As our dataset is imbalanced, we report the precision, recall, and F1-score (harmonic mean of the precision and recall) as shown in Table \ref{tab:themes_scores}. 


We observe that the classification of \textit{Empathetic Language} is particularly difficult. The best model is the transformer BERT model which achieves an F-1 score of 0.58. On the other hand, sentences with \textit{Medical Procedural Information} are much easier to identify with most classifiers achieving an F-1 score above 0.65. Sentences labeled \textit{Both} are increasingly difficult (best classifier score of 0.6 F-1). Classification scores for sentences containing \textit{Neither} fall just short of scores from \textit{Medical Procedural Information} sentences. 

\begin{figure}[ht]
    \centering
    \small
    \includegraphics[scale=0.35]{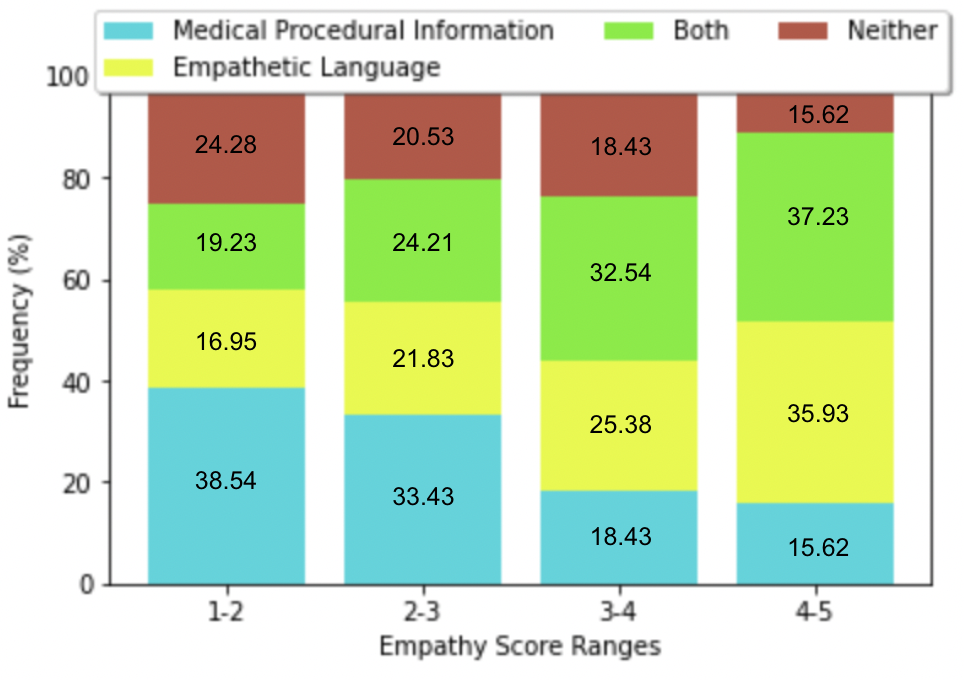}
    \caption{Frequency distribution (\%) of themes in essays for various empathy score ranges} 
    \label{fig:themes}
\end{figure}

To better understand how these themes correlate with the overall empathy score at essay level, we compare frequencies and distribution of each theme for various essay empathy score ranges (Figure \ref{fig:themes}) across the entire dataset. High empathy essays (scores >3) tend to show a large amount of \textit{Empathetic Language} and \textit{Both}, while low empathy essays (scores < 3) seem to favor \textit{Medical Procedural Information} language.

\begin{figure*}
    \centering
    \small
    \includegraphics[scale=0.35]{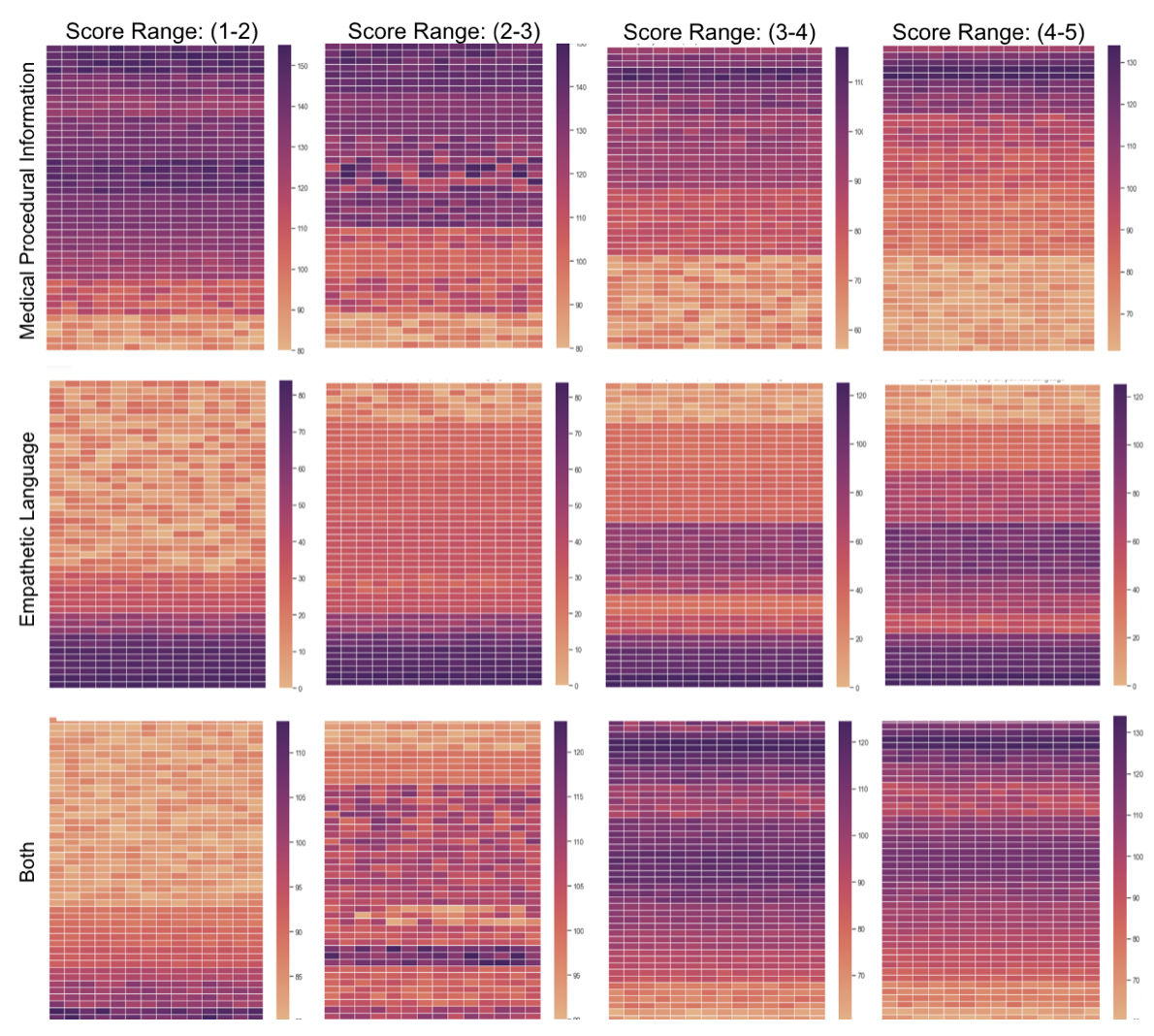}
    \caption{Heatmaps for themes in sentences of narrative essays across all overall empathy score ranges: Row\#1 shows heatmaps for \textit{Medical Procedural Information}; Row\#2 for \textit{Empathetic Language}; Row\#3 for \textit{Both}. Dark colors (purple) indicate that many essays exhibit the theme in the respective position of the essay. Light colors (yellow) indicate a small number of essays have occurrences of the theme for the given position.}
    \label{fig:heatmaps}
\end{figure*}

\vspace{0.05in}
\noindent
{\bf Heatmaps of Medical Narrative Essays}.
It is also interesting to visually analyze the distribution of these themes in the layout of the narrative essays. Thus, for each essay, we highlight the sentences containing each theme and generate heat maps that might highlight high theme concentrations. We standardized the format of each essay to an A4 paper,\footnote{Times New Roman, size 12: 42 lines of 14 words each} generating a 42 x 14 matrix. \footnote{We generated a separate heatmap (size: 81 x 14) for 24 essays since these were much longer and didn't fit on a standard A4 paper. These showed similar position patterns.} For each essay and position -- i.e., (row, column) -- we note the occurrence of each theme. We then build a heat map from these counts, thus generating 3 heatmaps, one for each theme along the following overall empathy score ranges: (1-2), (2-3), (3-4), and (4-5) (Figure \ref{fig:heatmaps}). 

The heatmaps for theme \textit{Medical Procedural Information} for low empathy score essays show darker colors (purple) indicating a higher frequency of use at the beginning and middle of the essay. Lighter colors (orange and yellow) showcasing lower concentrations of the theme seems to be more prevalent in higher empathy score essays. 
\textit{Empathetic Language} tends to increase in coverage (i.e., darker color portions) from low to high-score empathy essays, with a preference toward the end of the essay.\footnote{A  closer look indicates that students who wrote low-empathy essays showed a tendency to use some emotional language in the last paragraph - which appeared rather rushed and forced.}
\textit{Both} themes seem to concentrate, specifically towards the top and middle of the essays for high empathy scores (darker colors). Low empathy essays also show some shades of purple (i.e. some concentration) towards the bottom and lower third of the essays. 

\subsection{Incorporating Halliday Features into the Theme Classifier} 

In this section, we seek to improve our sentence theme classifier by incorporating the constructions and stylistic features identified in Section 4. For each sentence, we append a Boolean value indicating whether each feature is present in the given sentence -- e.g., if a sentence is in active voice (feature \textit{Active} is 1; feature \textit{Passive} is 0); if the sentence contains a HA+P (feature value is 1), and so on. Since in our baseline experiments the BERT model gave the best results across all 4 themes, we extend it here with all the features (construction-BERT) and report new scores (see bottom part of Table \ref{tab:themes_scores}). Indeed, the inclusion of these features yields better performance, with a large increase for most of our themes including, \textit{Empathetic Language}, \textit{Both}, and \textit{Neither}, and smaller performance increases in \textit{Medical Procedural Information}.

Leave-one-out feature contribution experiments (see bottom of Table \ref{tab:themes_scores}) show that removing \textit{Voice: Active} and \textit{Voice: Passive} slightly decreases performance in \textit{Empathetic Language} and \textit{Both} (with \textit{Voice: Active} providing the highest decrease).

Removing \textit{Processes} also shows a fair decrease in all themes except \textit{Neither} which shows no change in performance. A deeper analysis indicates that \textit{Processes: Material} helps with \textit{Medical Procedural Information} but hurts performance on \textit{Empathetic Language}. 

The constructions \textit{HA+P} and \textit{BP+P} are most important for classification; the removal of \textit{BP+P} yields the lowest F-1 score measure for detecting empathy. This shows the doctor (i.e., the student writer) paid particular attention to the patient's emotional state (thus showing empathy). Body parts in this type of discourse are particularly associated with non-verbal emotional language, which is highly indicative of empathy.
\textit{HA+P} is also an important feature for the theme \textit{Neither}. Removal of \textit{IE+P} gives a slight decrease in performance, while \textit{G+P} has almost no effect on the classification results.
Finally, the \textit{Tone: Energetic} and \textit{Tone: Static} features (constructionBERT-\textit{Tone}) show to be important for the themes \textit{Medical Procedural Information}, \textit{Empathetic Language}, and \textit{Both}. For \textit{Tone: Energetic}, there is a 0.02 decrease in F-1 for medical procedural information, and a 0.05 for \textit{Empathetic Language} and \textit{Both}. For \textit{Tone: Static}, we observe a decrease in performance for \textit{Empathetic Language} by 0.02 and \textit{Both} by 0.01. 

With our binary classification task, we see similar patterns as constructionBERT-Tone yields much lower performances. The energetic and static tones yield 0.004 and 0.01 increases in F-1 scores for \textit{Medical Procedural Information} and \textit{Empathetic Language}. Our analysis also showed that G+P (Goal+Process), Processes (Mental and Material), and HA+P (Human Actor+Process) were also increasingly important for score improvements.

Interested in directly comparing the \textit{Medical Procedural Information} and \textit{Empathetic Language} sentences, we further built a binary version of the simple BERT model, and another of constructionBERT, and found these tasks to be slightly easier. The binary BERT model achieved an F-1 score of 0.75 for \textit{Medical Procedural Information} and a 0.62 for \textit{Empathetic Language}. After adding the generated features (i.e., the binary constructionBERT), we see a small increase in F-1 scores (+0.01 for \textit{Medical Procedural Information} and +0.03 for \textit{Empathetic Language}).

Overall, the results of the effects of transitivity features on meaning, perceived agency and involvement of the Agent are in line with those obtained for literary genre texts by \citet{nuttall2019transitivity} through manual inspection. More specifically, the stylistic choices given by such linguistic constructions seem to be good indicators of the degree of perceived agency an Agent has in relation to others and the environment, as tested here for the empathy task on our dataset. In research on stylistics, the set and usage of such stylistic constructions and features in a text is known as the stylistic profile of the text. Encouraged by the correlations between Halliday's features with our essay level empathy scores, we would like to extrapolate and maintain that a set of rich stylistic constructions (like those tested in this research) can ultimately lead to informative \textbf{Empathy Profiles} -- essay level form-meaning-style structures that can give an indication of the degree of social and empathetic detachment of the doctor toward the patient. Of course, while more research is needed in this direction, we believe we showed here the potential of such an approach to the task of empathy detection classification overall, and to clinical context in particular.

\section{Conclusions}

Medical education incorporates guided self-reflective practices that show how important it is for students to develop an awareness of the emotional and relational aspects of the clinical encounter with their patients \cite{warmington2019storytelling}. The way people identify themselves and perform in particular roles and in relation to others brings together a specific set of values, attitudes,  and competencies that can be supported through ongoing self-reflection. Such interactions can be captured in language via constructions as part of CxG and Halliday's transitivity system. 

In this paper, we bring various aspects of these theories in a deep learning computational framework to model empathetic language 
in a corpus of essays written by premed students as narrated simulated patient–doctor interactions. 
We start with baseline classifiers (state-of-the-art recurrent neural networks and transformer models). Then, we enrich these models with a set of linguistic constructions proving the importance of this novel approach to the task of empathy classification for this dataset. Our results indicate the potential of such constructions to contribute to the overall empathy profile of first-person narrative essays.

\bibliography{anthology,custom}

\section{Appendix}
\label{sec:appendix}

\begin{figure}[H]
\centering
\begin{subfigure}{.5\textwidth}
  \centering
  \includegraphics[width=1\linewidth]{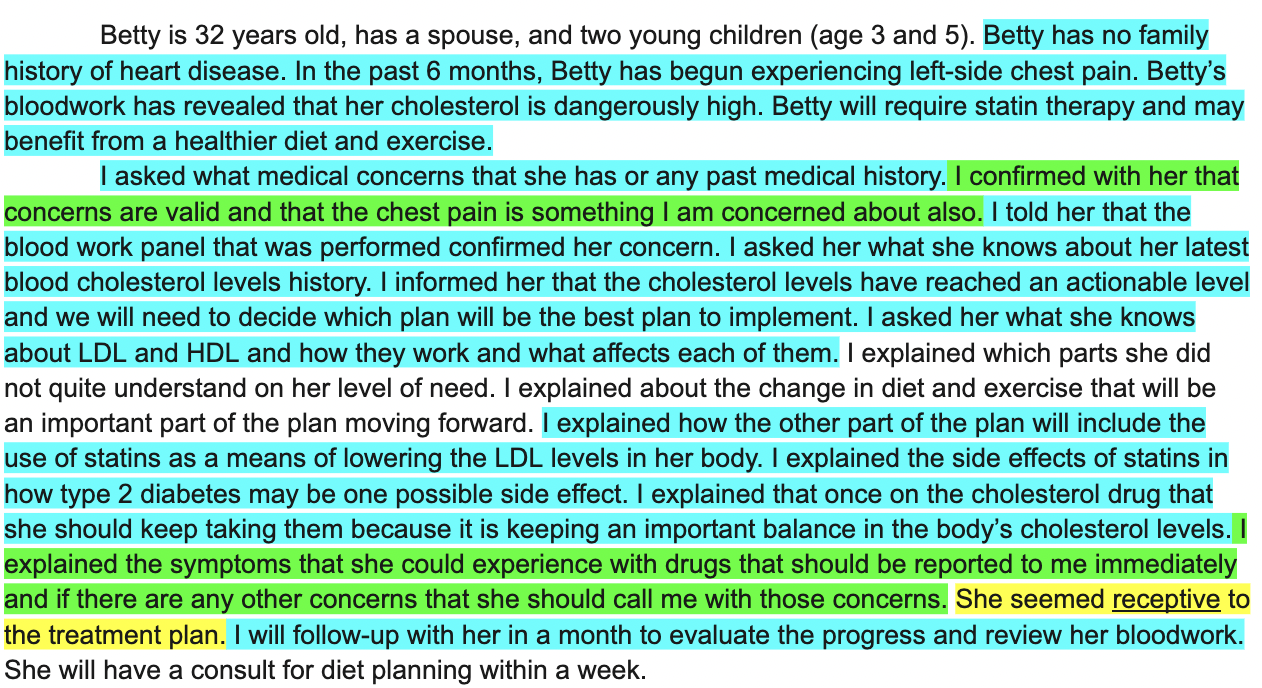}
  \caption{Example of Essay with Empathy Score: 1}
  \label{fig:sub1}
\end{subfigure}%
\begin{subfigure}{.5\textwidth}
  \centering
  \includegraphics[width=0.9\linewidth]{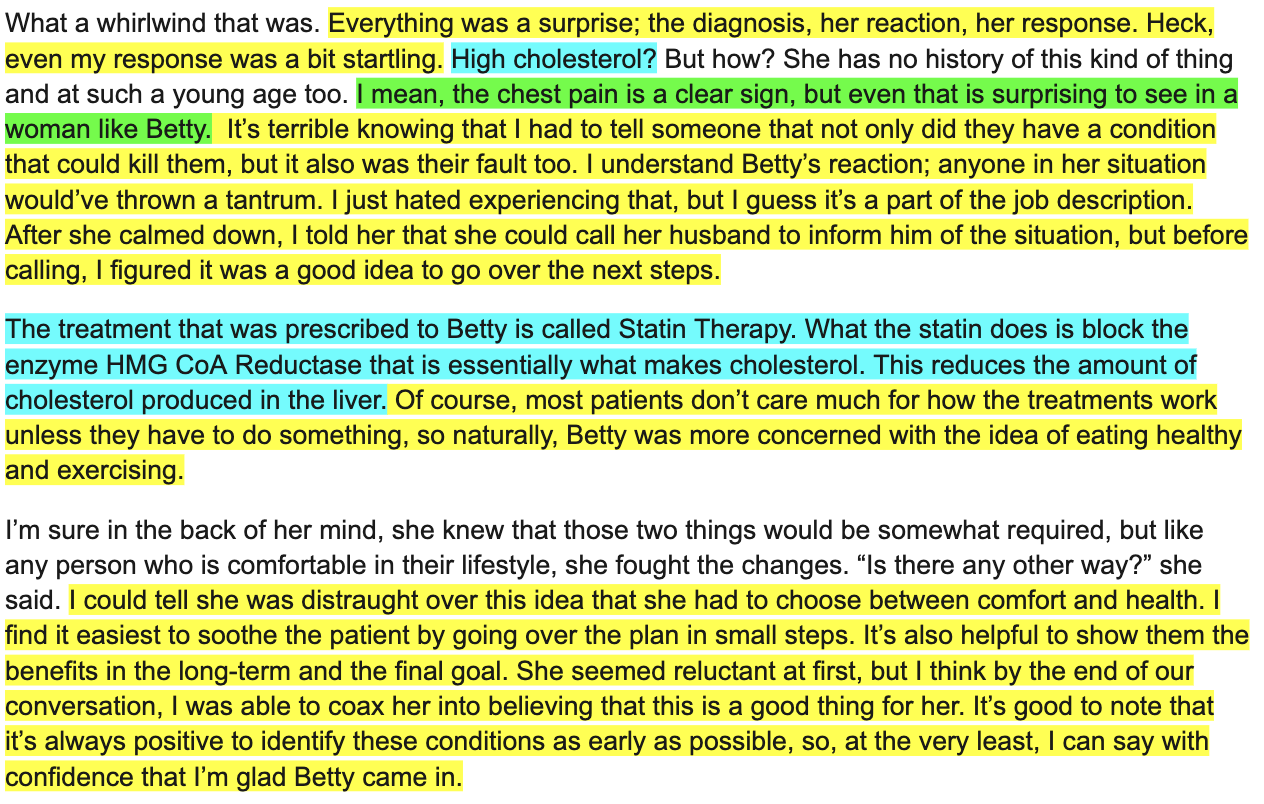}
  \caption{Example of Essay with Empathy Score: 5}
  \label{fig:sub2}
\end{subfigure}
\caption{Two Sample Essays from the Dataset Highlighted by Sentence Themes}
\label{fig:test}
\end{figure}

Figure \ref{fig:test}  shows two examples of essays, one with low empathy and one with high empathy,  highlighted with the themes: \emph{Medical Procedural Information} (cyan), \emph{Empathetic Language} (yellow), and \emph{Both} (green). \emph{Neither} sentences are not highlighted. It is interesting to see that in Essay (a), the sentences mentioning diet and exercise were not identified as \emph{Medical Procedural Information} given that they were not found in Dr. Kavita Ganesan’s work on clinical concepts \cite{Ganesan2016}.

\end{document}